\documentclass{article}
\usepackage{times}
\usepackage{graphicx}
\usepackage{latexsym}
\usepackage{natbib}

\begin{document}

\title{What Is It Like to Be a Brain Simulation?}
\author{Eray \"{O}zkural}

\maketitle
\bibliographystyle{plainnat}

\abstract{
We frame the question of what kind of subjective experience 
a brain simulation would have in contrast to a biological brain. We
discuss the brain prosthesis thought experiment. Then, we
identify finer questions relating to the original inquiry, and set out
to answer them moving forward from both a general physicalist
perspective, and pan-experientialism. We propose that the brain
simulation is likely to have subjective experience, however, it may differ
significantly from human experience. Additionally, we discuss the
relevance of quantum properties, digital physics, theory of
relativity, and information theory to the question.
}

\section{Introduction}
The nature of experience is one of those deep philosophical questions
which philosophers and scientists alike have not been able to reach a
consensus on. In this article, I review a computational variant of a
basic question of \emph{subjectivity}. In his classical article "What
is it like to be a bat?", Thomas Nagel investigates whether we can
give a satisfactory answer to the question in the title of his
article, and due to what he thinks to be fundamental barriers,
concludes that it is not something we humans can know
\cite{nagel-bat}. We can intuitively agree that although the bat's
brain must have many similarities to a human's, since both species are
mammalian, the bat brain contains a sensory modality quite unlike any
which we possess. By induction, we can guess that perhaps the
difference between sonar perception and our visual experience 
could be as much as the difference
between our visual and auditory perception. Yet, in some sense sonar
is both visual and auditory, and still it is neither visual nor
auditory. It is more similar to vision, because it helps build a model
of the scene around us, however, instead of stereoscopic vision, the
bat sonar can make accurate 3-D models of the environment from a
particular point of view, in contrast with normal vision that is said
to have "2-1/2D vision". Therefore, it is unlike anything that humans
experience, and perhaps our wildest imaginations of bat sonar
experience are doomed to fall short of the real thing. Namely, because
it is difficult for us to understand the experience of a detailed and
perhaps rapidly updated 3-D scene that does not contain optical
experience as there is no 2-D image data from eyes to be
interpreted. This would likely require specialized neural
circuitry. And despite what Nagel has in mind, it seems theoretically
possible to "download" bat sonar circuitry into a human brain (by
growing the required neural module according to a given specification,
connected to sonar equipment implanted in the body) so that
the human can experience the same sensory modality. In this problem,
armchair philosophy alone may not be sufficient. The
 barrier to knowing what it is like to be a bat is, thus, mostly a
technological barrier, rather than a conceptual or fundamental
barrier. 
Though, ultimately, it may be argued that we cannot expect 
one to know \emph{exactly} what a bat experiences, short of being
one. In the best case, we would know what a bat experience is like, as
the human brain could be augmented with a reconstruction of the
perceptual machinery.

That being the case, we may also consider what a brain simulation, or
an ``upload'' as affectionately called in science fiction literature,  would
experience, or whether it would experience anything at all, as brain
simulation is a primary research goal on which computational
neuroscientists have already made progress, e.g.,
\cite{Izhikevich_Edelman_2008}.
 The question that I pose
is harder because the so-called upload usually does not run on a biological
nervous system, and it is easier because the processing is the
simulation of a human brain (and not something else). Answering this
question is important, because presumably the (subjective) experience,
the raw sensations and feelings of a functional human brain, are very
personal and valuable to human beings. We would like to know, if there
is a substantial loss or difference in the quality of experience for
our minds' digital progeny.

A recent survey of large-scale brain simulation projects may be found
in \cite{degaris-sim}.

\section{Brain prosthesis thought experiment}

The question is also very similar to the brain prosthesis thought
experiment, in which biological neurons of a brain are gradually
replaced by functionally equivalent (same input/output behavior) synthetic
(electronic) neurons \cite{moravec-mindchildren}. In that thought
experiment, we ponder how the subjective experience of the brain would change. 
Although there are 
challenging problems such as interfacing smoothly with existing neural
tissue, it is a scientifically
plausible thought experiment, also discussed at some length in \cite[Section 26.4]{aima}. 
 Moravec suggests that
nothing would change with respect to conscious experience
in his book. Marvin Minsky has written similarly while discussing
whether machines can be conscious \cite{minsky-consciousmachines}.  
He produces an argument similar to Wittgenstein's ``beetle-in-a-box''
thought experiment: since a brain simulation is supposed to be
functionally equivalent, its utterances would be complete, and the
brain simulation would know consciousness and claim to be conscious;
why should we think that the simulation is lying deliberately? This is
a quite convincing argument, however, it only neglects to mention that
a form of \emph{physical} epiphenomenalism could be true (in that
experiential states may mirror electrical signals, yet they may be 
distinct physically).

Contrariwise,
John R. Searle maintains that the experience would gradually vanish in his
book titled ``The Rediscovery of the Mind'' \cite{searle-book}. 
The reasoning of Minsky and Moravec seems
to be that it is sufficient for the entire neural computation to be
 equivalent at the level of electrical signaling (as the synthetic
neurons are electronic), while they seem to disregard other brain
states. While for Searle, experience can only exist in "the right
stuff", which he seems to be taking as biological substrate, although
one cannot be certain \cite{searle-minds}. We will revisit this
division of views soon enough, for we shall identify yet another
possibility.

\section{Naturalist theories of experience}

An often underrated theory of experience is panpsychism, the view that
all matter has mental  properties.
 It is falsely believed by some that
panpsychism is necessarily incompatible
with physicalism. 
However, this is far from a settled controversy. 
Strawson has recently claimed that physicalism \emph{entails} panpsychism
\cite{Strawson2006}. We must also mention the view of
pan-experientialism: that experience resides
in every physical system, however, not everything is a conscious mind,
for that requires \emph{cognition} in addition. 
Panpsychism is also proposed as an admissible philosophical 
interpretation of human-like AI experience in
\cite{goertzel-hyperset}.



Why is this point of view (panpsychism) significant? The evidence from psychedelic
drugs and anesthesia imply that changing the brain chemistry also
modulates experience. If the experience changes, what can this be
attributed to? Does the basic computation change, or are chemical
interactions actually part of human experience? It seems that
answering that sort of question is critical to answering the question
posed in this article. However, it first starts with accepting that it
is natural, like a star, or a waterfall. Only then can we begin to ask
questions with more distinctive power.

Over the years, I have observed that neuroscientists were often too shy
to ask these questions, as if these questions were dogma. Although no
neuroscientist would admit to such a thing, it makes one 
doubt if religious or superstitious pre-suppositions may have a
role in the apparent reluctance of neuroscientists to investigate this
fundamental  question in a rigorous way. One particular study may shed
light on the question \cite{schneidman-universality} which 
claims that the neural code forms the basis of experience, therefore
changes in neural
code (i.e., spike train, a spike train is the sequence of electrical signals that
travel down an axon), change experience. That's a very particular
claim, that can be perhaps one day proven in experiment. However, at
the present it seems like a hypothesis that we can work with, without
necessarily accepting it.

That is to say, we are going to analyze this matter in the framework
of naturalism, without ever resorting to skyhooks. We can consider a
hypothesis as the aforementioned one, however, we will try to distinguish finely
between what we \emph{do} know and what is
\emph{hypothetical}. Following this methodology, and a bit of common
sense, I think we  can derive some scientifically plausible
speculations, following the terminology of Carl Sagan.

\section{The debate}

Let us now frame the debate more thoroughly, given our small
excursion to the origin of the thought experiment. 
On one side, AI researchers like Minsky and Moravec seem
to think that simulating a brain will just work, and experience will be
unchanged relative to the original brain. 
On the other side, skeptics like Searle and Penrose, try
everything to deny "consciousness" to poor machinekind. Although both
Searle and Penrose are purportedly physicalists, they do not refrain
from seeking almost magical events to explain experience.


However, it is not likely that word play will aid us much.
We need to have a good scientific theory of when and how
experience occurs. The best theory will have to be induced from
experimental neuroscience and related facts. What is the most basic
criterion for assessing whether the theory of experience is
scientifically sound? No doubt, it comes down to rejecting each
and any kind of supernatural/superstitious explanation and approach this
matter the same way as we are investigating problems in molecular
biology, that the experience is ultimately made up of physical
resources and interactions, and there is nothing else to it; this is a view
also held by Minsky as he likens mysticism regarding consciousness
to vitalism \cite{minsky-consciousmachines}. In
philosophy, this approach to mind is called physicalism. A popular
statement of physicalism is \emph{token physicalism}: every
mental state x is identical to a physical state y. That is a general 
hypothesis that neuroscientists already accept, because presumably, when the
neuroscientist introduces a change to the brain, he would like to see
a corresponding change in the mental state. One may think of
cybernetic eye implants and transcranial magnetic stimulation and
confirm that this holds in practice, and that the hypothesis need not be
questioned, for one would find it very difficult to find counter-examples.

\section{Asking the question in the right way}

We have discussed every basic concept to frame the question in a way akin
to analysis. Mental states are physical states. The brain states in a
human \emph{constitute} its subjective experience. The question is
whether a particular whole brain simulation, will have experience, and
if it does, how similar this experience is to the experience of a
human being. If the proponents of pan-experientialism are right, then
this is nothing special, it is a basic capability of every physical
resource (per the scientifically plausible, physicalist variant 
of pan-experientialism). 
However, we may question what physical states are part of human experience. We
do not usually think that, for instance, the mitochondrial functions
inside neurons, or the DNA, is part of the experience of the nervous
system, because they do not seem to be directly
participating in the main function of the nervous system:
thinking. They are not part of the causal picture of thought. 
Likewise, we do not assume that the power supply is
part of the computation in a computer.

This analogy might seem out of place, initially. If
pan-experientialists are right, experience is one of the basic features of the
universe. It is then all around us, however, most of it is \emph{not}
organized as an intelligent mechanism, and 
therefore, correctly, we do not call them conscious. This
is the simplest possible explanation of experience that has not been disproven
by experiment, therefore it is a likely scientific hypothesis. It does
not require any special or strange posits, merely physical resources 
organized in the right way so as to yield
an intelligent functional mind.  Consider my ``evil alien'' thought
experiment. If tonight, an
evil alien arrived and during your sleep shuffled all the connections
in your brain randomly, would you still be intelligent? Very unlikely, since
the connection pattern determines your brain function. You would most
likely lose all of your cognition, intelligence and memory. 
 However, one is forced to accept
that even in that state, one would likely have an experience, an
experience that is probably \emph{meaningless} and \emph{chaotic}, but
an experience nonetheless. Perhaps, that is what a glob of plasma
experiences. The evil alien thought experiment supports the
distinction between experience and consciousness. Many philosophers
mistakenly think that consciousness consists in experience. That, when
we understand the ``mystery'' of experience, we will understand
consciousness. However, this is not the case. Experience is part of
human-like consciousness, indeed, however, consciousness also includes 
a number of high-level cognitive functions such as reasoning,
prediction, perception, awareness, self-reflection and so forth
\cite[Section 4]{em}. 
I suggest that it is possible that there are minds without human-like
consciousness and with experience (e.g., a special purpose neural
network adding numbers), and experience without any recognizable
mentality.  

\section{Neural code vs. neural states}

Consider the hypothesis that experience is determined
by particular neural codes. If that is true, even the experience of
two humans is very different, because it has been shown that neural
codes evolve in different ways \cite{schneidman-universality}. 
One cannot simply substitute the code
from another human in someone else's brain, it will be random to the second
human. And if the hypothesis is true, it will be another kind of
experience, which basically means that the blue that I experience is
different from the blue that you experience, while we presently think
we have no way of directly comparing them. Strange as that may sound, as it is
based on sound neuroscience research, it is a point of view we must
take seriously.

Yet even if the experiences of two humans can be very different, they
must be sharing some basic quality or property of experience. Where
does that come from? If experience is this complicated time evolution
of electro-chemical signals, then it is in the shared nature of these
electro-chemical signals (and processing) that provides the shared
computational platform. Remember that a change in the neural code
(spike train) implies a lot of changes. First of all, the chemical
transmission across chemical synapses would change. Therefore, even a brain
prosthesis device that simulates all the electrical signaling
extremely accurately, might still miss part of the experience, if the
bio-chemical events that occur in the brain are part of
experience. Second, the electro-magnetic fields would
change. Third, the computation would change, although the basic
``program'' of the nervous system does not change from individual to
individual (neural processing does have many invariant properties,
such as coding efficiency). 

To answer the question decisively, we must first
encourage the neuroscientists to attack the problem of human
experience, and find the sufficient and necessary conditions for 
experience to occur, or be transplanted from one person to the
other. They should also find to what extent chemical reactions (or
other physical events) are
significant for experience. 

If, for instance, we find that the
distinctive properties of nervous system experience \emph{crucially} depend 
on quantum computations
carried out at synapses and inside neurons, that might mean that to
construct the same kind of experience you would need similar material
and method of computation rather than a conventional electronic computer.
 Quantum computation exploits the quantum
mechanical properties of matter, most notably superposition and coherence, to
achieve more efficient computation than classical computation which
does not use any quantum mechanical effects. 
 The basic unit of storage in a quantum computer, qubit,
relies on a quantum property known as quantum superposition, which
means that all theoretical states of a quantum system exist
simultaneously until measured. 
A qubit state is in superposition of $0$ and $1$ states,
representing both states simultaneously, $n$ qubits are in superposition
of $2^n$ n-bit states. Superposition of a quantum system lasts until
quantum decoherence, i.e., when a measurement is made 
\cite{Schlosshauer2005Decoherence}, and the negation of that,  the 
property of quantum states influencing each other in superposition,
free from environmental interference, is
called quantum coherence. Quantum computation operations transform all
the states in superposition at the same time, 
exploiting quantum parallelism, which
on particular probabilistic problems surpass the efficiency of classical
computation \cite{Deutsch1985Quantum}. The processing of qubits
must be achieved while preserving quantum coherence, otherwise the
quantum system turns into a classical system.
A good example of another macro-scale quantum
mechanical property is superconductance, which is conductance with
zero resistance, and it depends on the superconductor material to
exhibit quantum coherence. Recent experiments suggest
that quantum coherence plays a key role in photosynthesis
\cite{Panitchayangkoon27122011},  therefore
we cannot rule out that quantum coherence might be a necessary aspect of
brain operation, and brain's subjective experience.

On the other hand, we need to consider the possibility that electrical
signals may be a crucial part of experience, due to the power and
information they encompass, so perhaps any electronic device has these
electron motion patterns that make up most of your subjective experience. 
If that is true, the electronic devices presently would be
assumed to contain brain-like experience, for instance. Then, the
precise geometry and connectivity of the electronic circuit could be
significant. However, it seems to me that chemical states are just as
important, and if as some people hypothesize quantum phenomena play
a role in the brain, it may even be possible that the quantum
descriptions may be relevant. That is to
say, we cannot rule out any such hypotheses a priori even if they
sound uncanny. Presently, however, there does not seem to be any evidence  
that quantum randomness, or quantum coherence plays a role in nervous
system. Another possibility is that it may be found that
electromagnetic fields generated in the brain are crucial to
experience, in which case the topology, amplitude, timing and other
properties of electrical signaling may be relevant, i.e., anything
that would change the electromagnetic field.

\section{Simulation and transcoding experience}

At this point, the reader might be wondering if the subject were not
simulation. Is the question like whether the simulation of rain is
wet? In some respects, it is, because obviously, the simulation of
wetness on a digital computer is not wet in the ordinary sense. \footnote{However, we  can invoke the concept of "universal quantum computer" from theory
 \cite{Deutsch1985Quantum,Lloyd:Universal},
and claim that a universal quantum computer would re-instate
wetness, relative to observers in the simulation, since a universal
quantum computer is supposed to be able to perfectly simulate any
finite quantum system. However, there would be only simulated wetness
relative to observers outside the quantum simulation.}

We may reconsider the question of experience of a brain simulation. 
We have a human brain A, a joyous lump of meat, and its
digitized form B, running on a digital computer. Will B's experience
be the same as A's, or different, or non-existent?

Up to now, if we accept the simplest theory of experience (that it
requires no special conditions to exist at all), then we conclude
that B will have \emph{some} experience, but since the physical material is
different, it will have a different \emph{texture} to it. Otherwise,
an \emph{accurate} simulation, by definition, stores the same
functional organization of cognitive constructs, like perception, memory,
prediction, reflexes, emotions without significant information loss,
and since the dreaded panpsychism may be considered possible, they might
give rise to an experience somewhat similar to the human brain, 
yet the computer program B, may be experiencing something else at 
the very lowest level. Simply because it is running on some future
 nanoprocessor instead of the brain, the physical states have become
 \emph{altogether} different, yet their relative relationship,
 i.e., the \emph{logical structure} of experience, is preserved. 

Let us try to present the idea more intuitively. The
brain is some kind of an analog/biological computer. A memorable analogy
is the transfer of a 35mm film to a digital format. Surely, many
critics have held that the digital format will be ultimately inferior,
and indeed the medium and method of information storage is altogether
different but the (film-free) digital
medium also has its affordances like being able to backup and copy
easily. In both kinds of film, the same information is stored, yet the
medium varies. Or maybe we can contrast an analog sound synthesizer with a
digital sound synthesizer. It is difficult to simulate an analog
synthesizer, but you can do it to some extent. However, the physical
make-up of an analog synthesizer and digital synthesizer are quite
different. Likewise, B's experience will have a different physical
\emph{texture} but its \emph{organization} can be similar, even if
the code of the simulation program of B will necessarily introduce
some physical difference (for instance neural signals can be
represented by a binary code rather than a temporal analog signal). 
Perhaps, the atoms and thus, the \emph{fabric} of B's experience will be
different altogether as they are made up of the physical instances of
computer code running on a digital computer. As improbable as it may
seem today, 
these simulated minds will be made up of live computer codes, so it would be
naive to expect that their nature will be the same as ours. They are
not human brains, they are bio-information based artificial
intelligences. In all
likelihood, our experience would necessarily involve a degree of
unimaginable features for them, as they are forced to simulate our
physical make-up in their own computational architecture. This brings
a degree of relative dissimilarity. And other physical
differences only amplify this difference.

Assuming the above explanation, therefore, when they are viewing the
same scene, both A and B will claim to be experiencing the scene as
they always did, and they will additionally claim that no change has
occurred since the non-destructive uploading operation went
successfully. This will be the case, because the state of experience
is more akin to the short-term memory \cite{minsky1980}, or RAM of
computers.  
 There is a complex
electro-chemical state that is held in memory with some effort, by
making the same synapses repeat firing consistently, so that more or
less the same physical state is maintained. This is what must be
happening when you remember something, a neural state that is somewhat
similar to when the event happened should be invoked. Since in B, the
texture has changed, the memory will be re-enacted in a different
texture, and therefore B will have no memory of what it used to feel
like being A.

Within the general framework of \emph{physicalism}, we can
claim that further significant changes will also influence
B's experience. For instance, it may be a different thing to work on
hardware with less communication latency. Or perhaps if the simulation
is running on a very different kind of architecture, then the physical
relations may change (such as time and geometry) and this may
influence B's state further. We can imagine this to be asking what
happens when we simulate a complex 3-D computer architecture on a 2-D
chip. We must maintain, however, that strict physicalism
leads us to reject the idea that no mental changes happen when
significant physical changes happen. If that were possible, then we
would have to reject the idea that mental states are identical to
physical states, which would be dualism.

Moreover, a precise answer seems to depend on a number of smaller
questions that we have little knowledge or certainty of. These
questions can be summarized as: 
\begin{enumerate}
\item What is the right level of
simulation for B to be functionally equivalent to A (i.e., working and 
responding to stimuli in the same manner)?
\item How
much does the biological medium contribute to experience? 
\item Does experience crucially
depend on any uncanny physics like quantum coherence? 
\end{enumerate}

I think that the right attitude to answering these finer
questions is again a strict adherence to naturalism. For instance, in
3, it may seem easier to also assume a semi-spiritualist
interpretation of Quantum Mechanics, and claim that the mind is a
mystical, unobservable, ineffable soul. That kind of reasoning will
merely help to stray away from scientific knowledge.

\subsection{General physicalist perspective}

 At this point, since we
do not have conclusive scientific evidence, this is merely guesswork,
and I shall give \emph{conservative} answers.

\emph{Question 1: }
 If certain
bio-chemical interactions are essential for the functions of emotions
and sensations (like pleasure), then not simulating them
adequately would result in a definite loss of functional accuracy. B
would not work the same way, behaviorally, as A. This is true even if spike trains
and changes in neural organization (plasticity) are simulated
accurately otherwise. It is also unknown whether we can simulate at a higher
level, for instance via Artificial Neural Networks, that have
abstracted the physiological characteristics altogether and just use
numbers and arrows to represent A, although recent brain simulation
work shows that this might be possible \cite{Izhikevich04032008}.
 It is important to know these so
that B does not lack some significant cognitive functions of A, such as
emotions. The right level of simulation seems to be 
at the level of molecular interactions which would at
least cover the differences among various neurotransmitters, and which
we can simulate on digital computers (perhaps imprecisely,
though). At least this would be necessary because we know that, for
instance, neurotransmitter levels and distribution influence behavior.
Thus, it would be prudent to be able to accurately simulate
the neurologically relevant biochemical states and dynamics in the
brain, without necessarily simulating genetics or cell operation. Recently,
there has been promising work on mapping the biochemistry of the brain.

\emph{Question 2: }
This is one
question that most people avoid answering because it is very difficult
to characterize. The most general characterizations may use
information theory or quantum information theory. However,
in general, we may say that we need an appropriate physical and
informational framework to answer this question in a satisfactory
manner. In the most general setting, we can claim that ultimately
low-level physical states must be part of experience, because
\emph{there is no alternative} \footnote{The only alternative would be
  dualism, which is unacceptable to a physicalist}.
For a general physicalist, accepting a strong form of physicalism
(that every mental event/property/predicate is exactly physical), 
it seems prudent to think that the biological medium contributes to
experience insofar as it influences \emph{computational} states
relevant to cognition. Thus, physicalism may force us 
to accept that the physical details of 
both electrical and chemical transmission of neural signals would be
significant. In other words, a good deal of neurophysics would be
included, there may be no simple answer as
pan-experientialists hope.

\emph{Question 3: }
Some opponents of
AI, most notably Penrose, have held that "consciousness" is due to
macroscopic quantum phenomena together with Hameroff
\cite{Hameroff:1996}, 
by which they try to explain "unity of
experience". While on the other hand, many philosophers of AI think
that the unity is an illusion \cite{minsky-consciousmachines}. 
Yet, the illusion is something to
explain, and it may well be that certain quantum interactions may be
necessary for experience to occur, much like superconductivity. This
again seems to be a scientific hypothesis, which can be tested. For a
physicalist, thus, this is an unsettled matter, open to future
research, however there is the possibility.

\subsection{Pan-experientialist perspective}

It is my hope that the reader appreciates that pan-experientialism is
indeed the simplest theory of experience that is consistent with our
observations, that every physical system may have the potential for experience,
regardless of the confusion that surrounds the theory. Assume that the
theory is right. Then,
when we ask a physicist to quantify that, she may want to measure the
energy, or the amount of computation or communication, or information
content, or heat, whichever works the best. 
A general characterization of experience such that
it would hold for any physical system, may be defined precisely, and
may be part of experiments. It would seem to me that the best
characterization then would use information theory, because 
experience would not matter if it did not contain any information. For
instance, an experience without any information could not contain any
pictures or words.  

I suggest that we use such methods to clarify these finer
questions. Assuming the physicalist version of pan-experientialism I may
attempt to refine the answers above. The first question is not
dependent on experience, it is rather a question of which processes
must be simulated for correct operation, so the answer does not change.

\emph{Question 2:}
 The biological medium seems to contribute at least as much as
required for  correct functionality (i.e., corresponding to neural
information processing and biochemical changes precisely), and at 
most all the information as present in the
biological biochemistry (i.e., precise cellular simulations), if we
subscribe to pan-experientialism.  These 
might be significant in addition to electrical signals. The latter
part might sound like a kind of vitalism but it is not, the cellular
``experience'' might simply constitute the low level texture of the collective
experience of neural cell assemblies. Presently, it is difficult to
say we know that no cellular details contribute to experience,
however, we would start with the simpler former hypothesis, because
there does not seem to be any good reason why we should 
experience the effects of parts of machinery that do not contribute to 
cognition (such as mitochondria), yet pan-experientialism would force
us to maintain this hypothesis, since if anything can have experience,
so can the cells, and cells do have internal information processing as
well, and neurons are physically part of neural information processing.

\emph{Question 3:} Not
necessarily. According to pan-experientialism, it may be claimed to be false,
since it would constrain minds to uncanny physics (and contradict with
the main hypothesis). If, for instance, quantum coherence is indeed
prevalent in the brain and provides much of the "virtual reality" of
the brain, then the pan-experientialist could argue that quantum coherence is
everywhere around us. Indeed, we may have a rather primitive
understanding of coherence/decoherence processes yet, as that is itself part of
the unsettled controversies in philosophy of physics. Another
possibility is the use of a generalist physical approach that is
compatible with pan-experientialism, such as relativity.

\subsection{Further concerns}

Other finer points of inquiry may as well be imagined, and they are
sorely needed, because our capability of analysis is limited when we
cannot ask the right questions. If we forever dabble with
superstitious questions, we will never have started answering our
question. Thus, I have tried to show how to approach
the main question, and tried to give novel answers from both a general
physicalist perspective and physicalist pan-experientialism.

\section{Physical basis and quantifying dissimilarity}

Of the sufficient and necessary physical conditions, I have naturally
spent some time exploring the possibilities. I think it is quite
likely that quantum interactions may be required for human experience
to have the same quality as an upload's, since biology seems inventive
in making use of any properties, more than we previously thought,
for macro bio-molecules have been shown
to have quantum behavior and quantum properties have been observed in
biochemistry, i.e., photosynthesis. Maybe, Penrose is right about quantum
coherence. However, specific experiments would have to be conducted to
demonstrate it. It is easy to see why computational states would have to evolve, but
not necessarily why they would have to depend on macro-scale quantum
states (since there are classical computers), and it is an open question what this says precisely of systems that do
not have any quantum coherence. Beyond Penrose, I think that the 
particular texture of \emph{our} experience may indeed depend on chemical
states, regardless of quantum coherence. If, additionally,   
the brain turned out to be a quantum-computer under our very noses,
that would be fantastic and we could then emulate the brain states
very well on artificial quantum computers. In this case, assuming that
the universal quantum computer itself has little overhead, the quantum
states of the upload could very closely resemble the original.

Other physical conditions can be imagined, as well. For instance,
digital physics provides a clear framework to discuss
experience. The psychological patterns would be cell patterns in the
universal cellular automata. A particular pattern may describe a
particular experience. Then, two patterns are similar to the extent
they are syntactically similar. Which would mean that, we still
cannot say that the upload's experience will be the same. It will
likely be quite different, as the patterns will vary considerably.

One of my nascent theories is the Relativistic Theory of Mind, 
which tries to explain subjectivity of experience with concepts from the
theory of relativity. The principle of relativity might
be a foundation for the inherent subjectivity of experience. 
From the general relativity
viewpoint, it makes sense that
different energy distributions have different experiences, since that is
all that there is. What else could change?
We may also consider that experience may depend on measurements,
and \emph{measurements change relatively}.
While this falls under ``uncanny physics''
category, we cannot rule it out. We start with the analytical observation that
all experience is subjective: there is no objective experience. The
experience is always relative to an observer. However, the
observer itself is not a unified entity, it is made up of components that
themselves make observations relative to other components, down to the
level of quanta, and each small observation incurs a delay due to the
speed of light. Therefore, the brain may now be viewed as a system of
microscopic subjective observers, that make observations
of others, and send messages to others.
 Neural communication thus connects these different
localities, physically.  In 
relativity, communication cannot occur spontaneously, therefore no
brain can work as a truly unified entity. However, communication
obviously builds a macroscopic experience, which may manifest itself
as a somewhat stable electromagnetic field which pervades the brain.
  It may be 
thought that the entire dynamics of the brain thus
may be \emph{interpreted} as a 
``virtual pocket universe'' space-time with energy and topology of its own, 
which is difficult to observe from outside,
however, occasionally there is communication with outside since it is
separate but accessible. This energy distribution is part of our
physical brains, but it is also the part that is intelligent (rather
than metabolic), so that we
realize it exists, as it continually infers information about itself
and the world.
According to this theory anything might
have a psychological experience if it also had intelligent machinery,
such as a brain simulation, however its experience would be
significantly different than an original brain, since its geometry and energy
distribution is likely to be quite different.
This theory eliminates biochemistry borders, and predicts that there
could be merging of experiences with high-bandwidth low-latency 
connections to machines or other nervous systems. Note that the
relativistic approach originates from Einstein's thesis of
separability, that physical events have locality in space-time.
 Thus, according to Einstein's philosophy of science, it would
be impossible to imagine a mind that transcends physical limits or
does not have a location (i.e., dualism), the
mind consists of physical events that happen within a space-time
boundary. 

A general description of the relative dissimilarity between two
cognitive systems can be captured by algorithmic information theory.  Here, the
dissimilarity might correspond to saying that the similarity between A's and B's states depends on the
amount of mutual algorithmic information of a complete physical
description of A, and the
physical description of B, i.e., how much information do the two
descriptions share? As a consequence, the dissimilarity between two
systems would contain the informational difference in the low-level
physical structures of A and B, together with the information of the
simulation program (not present in A at all), and architectural
differences, which could be quite a
bit if you compare nervous systems and electronic computer chips
running a simulation. It seems that this difference is not so insignificant
that it will not have an important contribution to experience. That is
one of the reasons why I think that the answer to the question is 
that the brain simulation will likely have a new kind of subjective
experience. There are many ways this may not be true, however, if,
for instance, the electromagnetic field is the body of experience,
then a neural prosthesis device that accurately duplicates spike
trains could reconstruct experience perfectly in principle.

\section{Notes on methodology and terminology}

Conducting thought experiments is very
important, but they should be taken with care so that the thought
experiment would be scientifically plausible, even though
it is very difficult or practically impossible to realize. For that
reason, per ordinary philosophical theories of "mind", I go no further
than neuro-physiological identity theory, which is a way of saying
that your mind is literally the events that happen in your
brain. Rather than being something else like a soul, a spirit, or a
ghost. The reader may have also noticed that I have not used the word
"qualia" because of its somewhat convoluted connotations. I did refer
to  the quality of experience, which is something you can think
about. In all the properties that can be distinguished in this fine
experience of having a mind, maybe some of them are luxurious even;
and that is precisely why I used the word "quality" rather than "qualia" or
"quale". I do not think that the notion that there is a family
resemblance among biological brains is a troubling notion. For all we
know, there should be some significant resemblance if we take 
physicalism seriously.

Please also note that the view presented here is entirely different
from Searle, who seemed to have a rather vitalist attitude towards the
problem of mind. According to him, the experience vanishes, because
it is not built from the right stuff, which seems to be the specific biochemistry
of the brain for him \cite{searle-minds}. Regardless of the possibility of an
artificial entity to have the same biochemistry, this is still
arbitrarily restrictive. Some may call this attitude carbon-chauivinism, however I 
think it is merely idolization of earth biology, as if it is above
everything else in the universe.

\section*{Acknowledgements}

Thanks to Ben Goertzel for his detailed comments that improved the
article considerably.

\bibliography{aiphil} 

\end{document}